\documentclass[10pt,twocolumn,letterpaper]{article}

\usepackage{wacv}              

\usepackage{amsfonts}
\usepackage{amsmath}
\usepackage{amssymb}
\usepackage{tikz}

\definecolor{wacvblue}{rgb}{0.21,0.49,0.74}
\usepackage[pagebackref,breaklinks,colorlinks,allcolors=wacvblue]{hyperref}

\def\wacvPaperID{2076} 
\def\confName{WACV}
\def\confYear{2026}

\title{SDT-\textit{6D}: Fully Sparse Depth-Transformer for Staged End-to-End \\ 6D Pose Estimation in Industrial Multi-View Bin Picking}

\author{Nico Leuze,~Maximilian Hoh,~Samed Do\u{g}an,~Nicolas R.-Peña,~Alfred Schoettl\\
Institute for Applications of Machine Learning and Intelligent Systems\\
University of Applied Science Munich, 80335 Munich, Germany\\
{\tt\small nico.leuze@hm.edu, \tt\small alfred.schoettl@hm.edu}
}

\begin{document}
\maketitle
\begin{abstract}
Accurately recovering 6D poses in densely packed industrial bin-picking environments remain a serious challenge, owing to occlusions, reflections, and textureless parts. We introduce a holistic depth-only 6D pose estimation approach that fuses multi-view depth maps into either a fine-grained 3D point cloud in its vanilla version, or a sparse Truncated Signed Distance Field (TSDF). At the core of our framework lies a staged heatmap mechanism that yields scene-adaptive attention priors across different resolutions, steering computation toward foreground regions, thus keeping memory requirements at high resolutions feasible. Along, we propose a density-aware sparse transformer block that dynamically attends to (self-) occlusions and the non-uniform distribution of 3D data. While sparse 3D approaches has proven effective for long-range perception, its potential in close-range robotic applications remains underexplored. Our framework operates fully sparse, enabling high-resolution volumetric representations to capture fine geometric details crucial for accurate pose estimation in clutter. Our method processes the entire scene integrally, predicting the 6D pose via a novel per-voxel voting strategy, allowing simultaneous pose predictions for an arbitrary number of target objects. We validate our method on the recently published IPD and MV-YCB multi-view datasets, demonstrating competitive performance in heavily cluttered industrial and household bin picking scenarios. 
\end{abstract}
    
\section{Introduction}
\label{sec:intro}

\begin{figure}
    \centering
    \begin{tikzpicture}

    \node[inner sep=0pt] (russell) at (0,0)
        {\includegraphics[trim={4.8cm 0cm 0 0cm},clip,width=.3\textwidth]{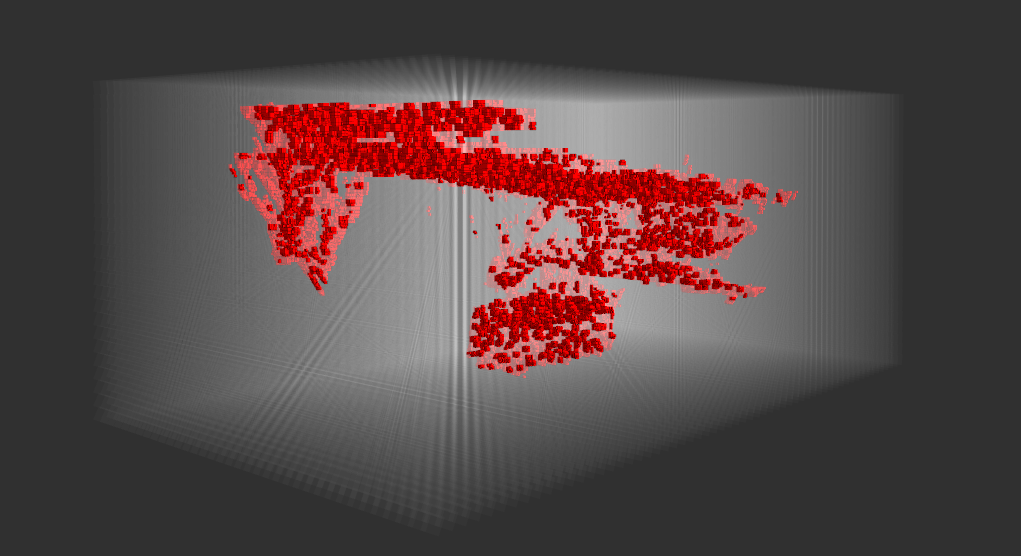}};
    \node[inner sep=0pt] (whitehead) at (4,0)
        {\includegraphics[width=.25\textwidth]{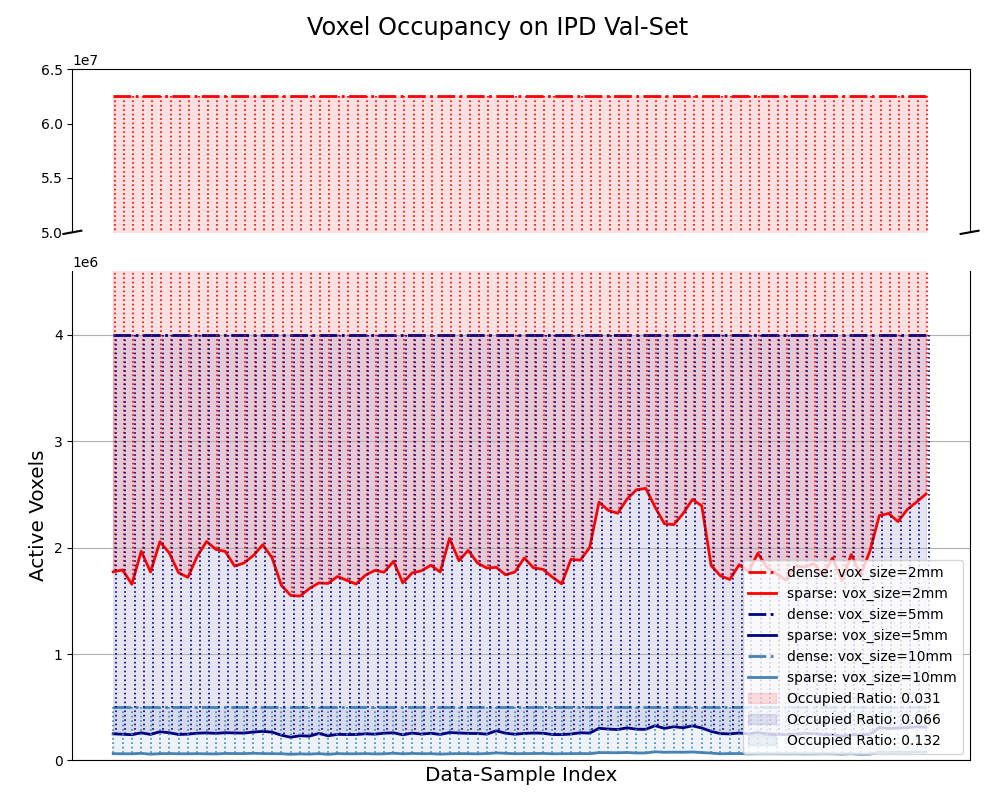}};
    \node[text width=4cm] at (1.4,-2.1) {\footnotesize{(a)}};
    \node[text width=4cm] at (5.9,-2.1) {\footnotesize{(b)}};
    \end{tikzpicture}
    \vspace{-0,8cm}
    \caption{Voxel occupancy statistics on the IPD dataset: (a) Occupied voxels (red) overlaid on the dense workspace grid illustrate the highly irregular and extremely sparse spatial distribution typical for bin-picking scenes (\(\vartheta=\) 8 mm). (b) Dense grids incur cubic complexity with increased resolution. Sparse voxel occupancy grows only squarely, enabling a more efficient 3D representation.} 
    \label{fig:voxel_occupancy}
    \vspace{-0,325 cm}
\end{figure}

\noindent Precise 6D pose estimation is foundational for robotic bin picking. Yet, especially industrial environments challenge vision systems with (self-) occlusion, clutter, and symmetric parts. Unlike well-textured objects, many industrial parts exhibit smooth, reflective surfaces and partial symmetries, which further confound traditional RGB- or depth-based detectors \cite{robi, symfm6d, posecnn, ffb6d, pvn3d, cozypose, sam6d}. Exploiting multi-view observations, however, have proven effective at alleviating these issues and deepen contextual awareness \cite{robi, symfm6d, mv6d, cozypose}. \\

\noindent While 3D representations offer a natural modality to the task of 6D pose estimation and have shown potential of leveraging geometric and topological properties more effectively \cite{pointnet, pointnet++, 3dssd, frustumpointnet, cloudaae, vgn}, the prohibitive memory and compute demands at high resolution often exceed feasible limits for deployment. Especially, conventional densely voxelized approaches impose unviable computation complexity, suffering from cubic scaling with resolution (\mbox{cf. Fig. \ref{fig:voxel_occupancy}(b)}) and thus forcing methods to operate at low resolutions \cite{vgn, deepsliding, activegrasp, voxelnet}. Bin-picking scenes exhibit extreme spatial sparsity at fine resolution scales. We observed that active occupancy rates fall to merely 3\% on the IPD dataset \cite{ipd} at reasonable resolution, rendering most computational effort on dense representations effectively meaningless. This underscores the potential of sparse 3D encodings, confining computation to active sensing regions. \\

\noindent Although inherently sparse, point-based detectors are constrained by time-consuming neighborhood queries for global context reasoning and inevitable information loss imposed by random downsampling \cite{pointnet, pointnet++, 3dssd}. Sparse voxel detectors, on the other hand, have recently become the dominant paradigm in long-range perception \cite{fstr, voxelnet, sphereformer, voxelnext}, offering state-of-the-art performance through scalable and efficient network architectures. By restricting computation to occupied voxels, these approaches naturally adapt to preserving scene sparsity. This yields favorable runtime and memory characteristics, while retaining high representational expressiveness and reasonable global context aggregation. \\

\noindent Yet, these potentials remain heavily underexplored in 6D pose estimation. Despite significant progress, recent methods for 6D pose estimation in cluttered bin-picking are constrained by the reliance on either (1) instance-level assumptions for accurate detections \cite{stablepose, megapose6d, cloudaae, cozypose, transpose} or (2) indiscriminate scene reduction strategies \cite{swindepose, pvn3d, ffb6d, mv6d, symfm6d}. The first line of research are multi-stage pipelines that sidesteps the core challenge of localizing hard-to-detect objects in clutter and inherits the weaknesses of upstream modules, as errors in segmentation or cropping directly propagate to pose estimation and severely limit robustness and adaptivity in cluttered scenes. Conversely,  holistic methods turn to random subsampling of images or point clouds to control computational costs. Such approaches, while practical, inevitably discard useful context and introduce variance in the preserved observation. Neither strategy adapts to the scene itself, which leads to information loss or inefficient use of capacity, leaving open the question of how to balance efficiency with fidelity in a scene-adaptive manner. \\

\noindent We answer this with a novel fully-sparse, depth-only approach to 6D pose estimation that fuses multi-view depth maps into either a fine-grained 3D point cloud, in its vanilla version, or a sparse TSDF. We hierarchically extract foreground regions via a staged heatmap strategy and incrementally lift the feature map resolution as data flow toward the pose prediction head. By building atop a fully sparse architecture, our approach dynamically constricts computational effort to the local point cloud density and the number of object candidates, yielding a streamlined end-to-end solution that seamlessly merges localization into 6D pose regression. Unlike prior sparse methods developed to cope with the vast observation space of long-range perception, our method uses sparsity not only for background suppression but also for preserving the sharp, high-resolution geometric detail essential for precise 6D pose estimation in cluttered robotic manipulation. Further, we propose a density-aware sparse transformer block that dynamically attends to (self-) occlusions and the non-uniform distribution of sparse 3D representations. In contrast to popular per-instance pipelines, our method processes the entire scene holistically via a novel per-voxel voting strategy, allowing simultaneous and dynamic pose prediction for an arbitrary number of target objects. We summarize our main contributions as follows. 

\begin{itemize}
    \item We rethink the architecture of current mainstream pose estimation approaches. With pilot experiments, we point out the potential of a staged heatmap core to allow holistic 6D pose regression at highest resolutions. Our proposed method eliminates the reliance on indiscriminate scene reduction heuristics or explicit per-instance assumptions.
    \item We propose a scene-adaptive heatmap strategy to pinpoint high-confidence foreground regions, drop background and preserve sample-level context. 
    \item Our method introduces a novel per-voxel voting mechanism that enables holistic scene reasoning in depth-only 6D pose estimation and supports scalable, adaptive pose prediction for an arbitrary number of objects.
\end{itemize}

\section{Related Works}
\label{sec:formatting}

\subsection{Depth-Based 6D Pose Estimation}
\noindent Although recent methods \cite{sam6d, foundationpose, freeze, mv6d} achieve strong results on household object datasets \cite{ycb, icbin, lm-o}, the estimation of 6D poses in industrial scenarios remains highly challenging due to clutter, occlusion, reflectivity, textureless objects and scene complexity. Due to the absence of reliant appearance features, 6D pose estimation for textureless objects is often addressed using depth data or RGB-D images \cite{mv-rgb25}. The most recent of these approaches to 6D pose estimation rely on multi-stage pipelines that assume pre-normalized object detections as input. CloudAAE \cite{cloudaae} stands out as a pioneering work that regresses instance-level 6D poses via an augmented autoencoder from 3D point clouds only. Yet, this approach relies on upstream detections that yield point cloud segments. The authors of \cite{stablepose} introduced StablePose that predicts 6D object poses based on geometrically stable patch groups from depth data. They also assume cropped point cloud segments as input. MegaPose \cite{megapose6d} assumes normalized RGB or RGB-D detections, and proposes a zero-shot framework to predict 6D object poses via a render and compare mechanism. TransPose \cite{transpose} introduces the concept of geometry-aware Transformers to the problem of cluttered 6D pose estimation. While assuming instance segmentation masks to extract instance-level point cloud crops, they employ a Graph-Convolutional feature extractor and a learnable position encoder as an upstream module to the transformer encoder. \\

\noindent Holistic methods \cite{ffb6d, pvn3d, swindepose} accept raw RGB-D observations and regress 6D poses on a sample-level. \cite{ffb6d} introduces a full flow bidirectional fusion approach which learns to combine appearance and geometry information for advanced representation learning. \cite{pvn3d} built upon two separate feature extractors for color and depth information, fuses the features into a dense map and feeds it to a Hough Voting 3D keypoint detection network. The depth-only SwinDePose \cite{swindepose} utilizes two encoder-decoder networks to learn representations from a 3D point cloud and vector angle image. The respective features are fused and fed into the semantic segmentation and keypoint detection modules, yielding instance-level object candidates. To control computational complexity, these methods rely on indiscriminate scene reduction strategies yielding inevitable information loss or redundant computation.  \\

\noindent Using a collection of RGB(-D) frames from distinct viewpoints, multi-view pose estimators mitigate occlusions and boost the reliability of 6D pose estimation. CozyPose \cite{cozypose} tackles 6D pose estimation using multiple RGB views, predicting object candidates per image with an upstream detector. The poses are refined through object-level bundle adjustment, while failing when an object appears in only one view. The authors of \cite{mv6d, symfm6d} fuse the geometrical and visual features from multiple RGB-D views before predicting the poses based on keypoints and least-squares fitting. Both methods employ learnable point-to-pixel fusion modules for pose regression, demanding fixed-sized scene reduction strategies. \cite{mv-rgb25} recently introduced an RGB multi-view method that relies on 2D detections and afterwards decouples the 6D pose estimation task into 3D translation regression via 2D vector fields and multi-view optimization followed by the rotation prediction through a max-mixture optimization. While multi-stage approaches fundamentally outsource the problem of localizing target objects in cluttered scenes, neither of the reviewed holistic (multi-view) methods adapt to the scene information itself.

\subsection{Sparse Detectors}
\noindent Sparse 3D detectors recently demonstrated remarkable performance in long-range perception. SECOND \cite{second} stands out as a pivotal work, addressing 3D object detection via a sparse backbone. The works of \cite{sphereformer, fstr, voxelnext, voxelnet, sst} adopted and advanced the idea of sparse processing up to fully sparse pipelines for better performance-efficiency trade-off. We argue that it is imperative to explore these potentials for cluttered 6D pose estimation. Although long-range 3D perception and (multi-view) bin-picking both face varying point density, spatial sparsity, and strong foreground–background imbalance, it is a non-trivial task to transfer design-choices from long-range perception to the cluttered 6D pose estimation task: Most recent fully sparse detectors regress the 4D pose (translation and z-rotation) from a single query voxel per object candidate via offset regression. This design heavily depends on distinctive feature exploration and local explicitness. This is effective in long-range perception where objects are large, spatially distinct, and comparatively well separated. In contrast, bin-picking scenes contain cluttered, highly similar, and small-scale objects whose boundaries frequently overlap. Here, sparse queries struggle to isolate individual items, causing unstable offset regression and degraded pose estimation performance. With our approach, we aim to resolve these conflicts while maintaining the core strengths in representation, efficiency and scalability.

\section{Method}


\begin{figure*}[!t]
  \centering
   \includegraphics[width=1.\linewidth]{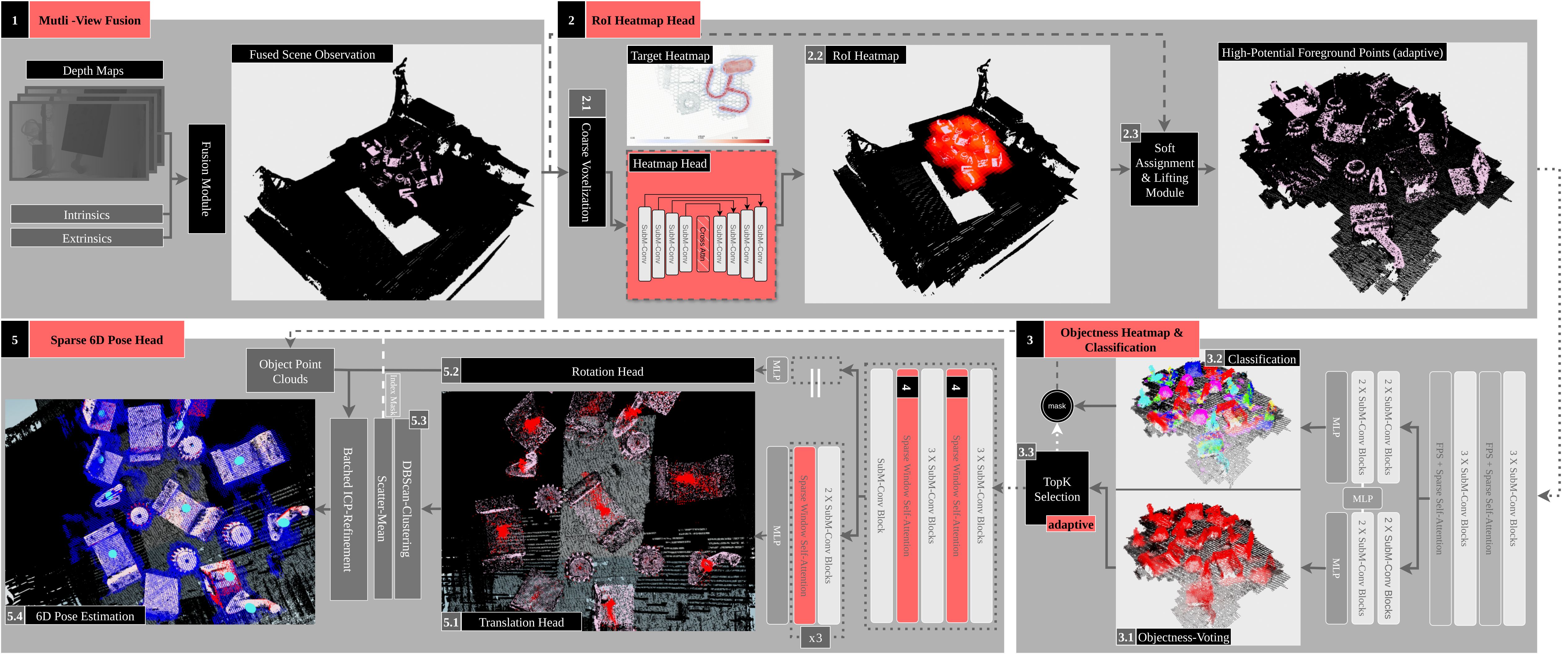}

   \caption{Overview of our framework architecture. Best viewed in color. \textbf{[1]:} Multiple raw depth observations are fused and discretized into a fine-grained sparse 3D voxel grid. \textbf{[2]:} The grid is more coarsely discretized (2.1) and fed into the \textbf{RoI Heatmap} that consists of a sparse U-Net. The most potential foreground voxels (visualized in ascending red) (2.2), along the global context features are lifted to the original high-resolution, while background voxels are dropped via soft assignment (2.3). \textbf{[3]:} The sparsified, yet high-resolutional, sparse grid is fed into the \textbf{Objectness Heatmap}. Fully sparse feature extraction layers are applied to obtain sharp objectness votes (3.1) and predict the per-voxel class (3.2). Based on the objectness scoring, we further sparsify the grid via an adaptive \(topK\) selector (3.3). \textbf{[5]:} The resulting voxels (depicted in blue in the bottom left image) represent the extremely sparse input to the \textbf{Sparse 6D Pose Head}. Sparse Convolutional Layers are interleaved with \textbf{Sparse Transformer Blocks} \textbf{[4]} to extract fine geometric details and local context. Relative translation offsets (5.1) and rotation estimations (5.2) are predicted on a voxel-level. The clusters, formed by the offset predictions, are used for instance indexing (5.3). Given the canonical object point clouds, we apply a batched ICP for pose refinement (5.4).}
   \label{fig:modelarch}
\end{figure*}

\noindent Within this chapter, we introduce the \textit{SDT-6D} framework. Tackling multi-view 6D pose estimation in cluttered robotic bin-picking scenarios, we propose a holistic approach that balances efficiency with fidelity in a scene-adaptive manner. To ensure broad adaptability, our method is designed to run efficiently on a single consumer-grade GPU.  The overall framework is shown in Fig.\ref{fig:modelarch} and contains five main parts: Multi-View Fusion (Sec. 3.2), RoI Heatmap (Sec. 3.3), Objectness Heatmap (Sec. 3.4), Sparse Transformer Block (Sec. 3.5) and the 6D Pose Regression (Sec. 3.6). 

\subsection{Overview}
Given multiple depth observations, multi-view 6D object pose estimation seeks to recover the rigid transformation \(T \in SE(3)\) that maps the object’s local coordinate frame to the reference world frame. This transformation comprises a rotation \(R \in SE(3)\) and a translation \(t \in \mathbb{R}^3\). We assume the respective camera's intrinsic and extrinsic parameters to be known. We tackle the problem of 6D pose estimation for known objects, demanding the 3D object point clouds. \\

\noindent Holistic 6D pose estimation in cluttered bin-picking faces an acute conflict between resolution and memory feasibility, particularly on deployable hardware. We aim to address this in a scalable and scene-adaptive manner. We fuse raw depth maps into a fine-grained, spatially sparse 3D voxel grid. To fully leverage sample-level information, we then apply a staged heatmap mechanism that progressively focuses the network on salient foreground voxels and discards confidently identified background in two main steps. (1) The Region of Interest (RoI) Heatmap Head, comprised of a submanifold sparse convolution \cite{second} backbone, takes a coarse representation of the sparse 3D voxel grid as input, extracts global context features and assigns per-voxel importance scores for pose estimation. Building upon the global context, we lift the features and importance scores to the initial fine-grained grid resolution and discard the high-confidential background voxels via soft assignment, while preserving sample-level information. (2) The resulting sparsified yet high-resolutional 3D grid is fed into the second heatmap to identify voxels (partially) occupied by target objects at reasonable resolution. Using an adaptive \(topK\) selector, we sparsify the grid to task-relevant 3D segments that preserve fine geometric details and local context information. This strategy enables levels of detail (up to \(\sim1\) mm resolution) otherwise unattainable under practical memory constraints. The sparse 6D head integrally regresses per-voxel estimates on the remaining grid (blue voxels in Fig. \ref{fig:modelarch}, bottom left image) for the relative translation offset and object rotation. Clustering, a scatter-operation and a pose refining ICP finally yield the 6D object pose estimations. Our method fundamentally departs from prior work, which is either detection-dependent at the instance level or applies random scene downsampling to fixed sizes, a limitation magnified in multi-view approaches. By exploiting low-resolution information as a scaffold for targeted high-resolution refinement, our method preserves global context without reverting to isolated object pipelines.  \\

\subsection{Sparse 3D Representation}
\noindent Let the continuous observation space be \(\mathcal{W} \subset \mathbb{R}^3\). We assume \(n\) depth maps \(\{\mathcal{D}_1, .., \mathcal{D}_n  \mid \mathcal{D}_i: \Omega_i \xrightarrow{} \mathbb{R}^+ \}\), each defining per-pixel depth values over the image domain \(\Omega\), with the corresponding camera intrinsics \(\{\mathcal{K}_1, .., \mathcal{K}_n\}\) and extrinsics \(\{\mathcal{T}_1, .., \mathcal{T}_n\}\). \\

\noindent \textbf{Vanilla Version.} We formulate the reconstruction problem as computing the globally consistent 3D point cloud \(\mathcal{P} = \bigcup_{i=1}^n \mathcal{P}_i \). Explicitly, we backproject each valid pixel \(\rho = [u,v]^{T} \in \Omega \) into 3D world coordinates with \(p_{i,j} = \mathcal{T}_i \hspace{0.1cm} (\mathcal{K}_i^{-1} \hspace{0.1cm} [u \hspace{0.15cm} v \hspace{0.15cm} 1]^T \hspace{0.1cm} \mathcal{D}_i(\rho)) \hspace{0.1cm} \in \mathbb{R}^3 \) and obtain \(\mathcal{P}_i = \{p_{i,1},..,p_{i,J}\}\). The fused point cloud \(\mathcal{P} \subset \mathbb{R}^{N \times 3}\) is given w.r.t. the reference frame.\\

\noindent \textbf{Sparse Truncated Signed Distance Field.} TSDFs offer volumetric expressiveness beyond surface boundaries, encoding both interior and exterior volumetric structures that can be leveraged for downstream robotics tasks such as antipodal grasp estimation \cite{vgn, activegrasp}. However, traditional dense representations incur prohibitive memory costs at scale. Hence, we build upon the work of \cite{o3d} to propose a sparse TSDF representation \(\phi: \mathbb{R}^3 \xrightarrow{} [-1, 1]\) as enriched input to the \textit{SDT-6D} framework. \\

\noindent We define a fixed, yet coarse, block size \(B > 0\) and discretize \(\mathcal{W}\) into axis-aligned blocks, indexed by \(b \in\mathbb{Z}^3\). Given the spatial localization property of voxels, let the spatial extent of a block be denoted as \(E_b \subset \mathbb{R}^3\).
Given an implicit surface observation \(\mathcal{P} = \{p_1, .., p_N\}\), obtained as for the vanilla version, we define an indicator \(I\), 
\begin{equation}
    I(b) \;=\;
  \begin{cases}
    1, & E_b \cap  \mathcal{P}\neq\emptyset,\\
    0, & \text{otherwise}.
  \end{cases}
\end{equation}
Let the set of active blocks be defined as \({\mathcal{B}=\{b : I(b)=1\}}\). Each spatial block \(E_b \in \mathcal{B}\) is subdivided into a regular, fine-grained \(L \times L \times L\) voxel grid \(V_{b} \) with voxel size \(\vartheta_{tsdf}\) and voxel index \(\ell \in \{0,..,L-1\}^3\). The active blocks with the flattened voxels are stored in a hashmap, so global memory grows in \(O(\mid \mathcal{B} \times L^3 \mid)\). The number of active voxels is \(M=B \; L^3\). For each active voxel center \(v_{b,\ell} \in V_{b}\), we project it into view \(i\) by \(\rho = \pi(\mathcal{K}_i\hspace{0.1cm}\mathcal{T}_i^{-1} \hspace{0.08cm} v_{b,\ell})\) and compute the signed distance at depth observation \(d_i = \mathcal{D}_i(\rho)\) as
\begin{equation}
    s_i(v_{b,\ell}) = d_i - z \;\;\;\;\;\; \text{with} \;\; (x,y,z) = \mathcal{T}_i^{-1} \; v_{b,\ell} \;\;\; .
\end{equation}
The normalized truncated SDF value is obtained with \(\phi_i \left( v_{b,\ell} \right) = \text{clamp}\left( s_i \left( v_{b,\ell}\right) / \tau_{\text{sdf}}, -1, 1 \right) \) where the truncation scalar \(\tau_{\text{sdf}} > 0\) is a constant. The signed distance values \(S = \{s_i \;: \; i=1,..,M\} \in [-1,1]\) are updated by weighted average across the views \cite{o3d}. From the set of active voxel centers we construct the 3D point cloud \(\mathcal{P}_{\text{sdf}} \subset \mathbb{R}^{M \times 3}\) and concatenate the normalized truncated signed distance values \(S_{\text{sdf}} \subset \mathbb{R}^{M\times1}\). We define the resulting sparse 3D representation as \(\Bar{\mathcal{P}} \subset \mathbb{R}^{M\times4}\). \\

\noindent Starting with any fused input representation, we construct an initial fine-grained sparse 3D voxel grid with voxel size \(\vartheta\). We label each voxel by \(x=(v,f)\) with index vector \({v \in \mathbb{Z}^3}\) and feature vector \(f \in \mathbb{R}^C\). The voxel grid is considered the attributed sparse voxel set \({\mathcal{X}=\{x_i : \; i=1,..,N \}}\), where \(N\) is the number of voxels. 


\subsection{RoI Heatmap Head}
\noindent In this section, we propose the lightweight, fully sparse RoI heatmap to identify high-confidential foreground regions and aggregate global context information on a sample-level. \\

\noindent To mitigate the quadratic scaling of memory costs, we construct a coarser version of the input sparse voxel set \(\mathcal{X}\). We use the voxel size \(\vartheta_{coarse} = 10 \; \vartheta\). The resulting grid \(\mathcal{X}'\) is the input to the RoI Head. For an accurate resolution-efficiency trade-off, the heatmap comprises a \mbox{U-Net} of residual submanifold convolution layers to overcome the sparse convolution dilation paradigm \cite{spconv}. Inspired by the work of \cite{voxelnext, fstr}, we apply Gaussian-smoothed heatmap targeting. We extend this strategy by jointly targeting both object centers and boundary voxels. Boundaries encode critical geometric structures. Centers may reside outside the implicit surface representation, thus requiring surrounding evidence for robust localization. As shown in \cite{voxelnext}, Gaussian smoothing helps reduce the impact of noise and forces the model to preserve local context for the downstream tasks. During model training, let \(C = \left(c_1, .., c_m\right)\) denote the ground truth centroids of the \(m\) target objects. Let the canonical object point clouds, transformed to the reference world frame, be further defined as \(\mathcal{O} = \left(O_1, .., O_m\right)\). \mbox{We score} a sparse voxel \(x' \in \mathcal{X}'\) with spatial position \mbox{vector \(p'\)}, using the distance weighting function 
\begin{equation}
    \begin{aligned}
    H(p') = \frac{1}{2} & \left( \exp\left(-\frac{\min_i\mid\mid p'-c_i \mid\mid^2}{\sigma_c^2}\right) + \right. \\ 
    & \left.\exp\left(-\frac{\min_{i,j}\mid\mid p'-O_{i,j} \mid\mid^2}{\sigma_b^2}\right) \right) \;\; ,
    \end{aligned}
\end{equation}
\noindent where \(H(p') \in [0,1]\) is the target score for sparse voxel \(x'\). \(\sigma_c\) and \(\sigma_b\) control the spread of influence for the object's center and boundaries, respectively.  \\

\noindent For sparse voxels \(x' \in \mathcal{X}'\), the proposed RoI Heatmap estimates per-voxel scores, which we denote as \(\hat{H} \in [0,1]\). Notably, the proposed targeting strategy leads a strong imbalance between foreground and background voxels. We tackle this by optimizing the heatmap head using the Gaussian Focal Loss \cite{cornernet} 
\begin{equation}
\mathcal{L}_{\text{RoI}} = 
- \, H \, (1 - \hat{H})^{\gamma} \log(\hat{H}) 
- (1 - H)^{\alpha} \, \hat{H}^{\gamma} \log(1 - \hat{H}) \;,
\end{equation}
\noindent where the hyperparameters \(\alpha\) and \(\gamma\) modulate the relative suppression of easy negatives (background) and the focusing on difficult voxels (foreground), respectively. \\

\noindent Given the inherent variability in foreground-to-background voxel ratios across 3D scenes \cite{sphereformer}, e.g. due to the number and spatial expression of target objects, we favor a sample-adaptive modulation scheme over fixed-sized \(topK\) selection. Thus, we apply soft suppression to the sparse voxels \(x'\). This promotes the preservation of only semantically salient foreground voxels and prevents redundant computation in a scene-adaptive manner. Given the per-voxel score \(\hat{h} \in \hat{H}\), we define the soft attention weight \(a_i\) as  
\begin{equation}
    a_i = \sigma \left( \beta \left( \hat{h}_i - \epsilon \right)\right) \in [0,1] \;\; ,
\end{equation}
\noindent with \(\beta >0\) controlling the sigmoid slope and a shift \mbox{offset \(\epsilon\)} to promote clear distinction. We drop highly confidential background voxels by thresholding the attention priors \(a_i\) at \(\kappa\), as \(\mathcal{X}'_{imp} = \{x' \in \mathcal{X}' \; \mid \; a_i > \kappa\}\). We frame the feature map lifting as the voxel grid search of \(\mathcal{X}\) in \(\mathcal{X}'_{imp}\), i.e. we select all sparse voxels \((v, f) \in \mathcal{X}\) that would be assigned to voxel grid indices \((v', f') \in \mathcal{X}'_{imp}\) at coarser resolution \(\vartheta_{coarse}\). We further enrich a selected voxel \(x\) with the feature of the matched voxel \(x'\), to preserve global context information. The resulting sparsified voxel grid is denoted as \(\mathcal{I}=\{x_1,..,x_{\mathcal{N}}\}\) (cf. upper right image in \mbox{Fig. \ref{fig:modelarch})}. Notably, the number of voxels \(\mathcal{N}\) is dynamic.


\subsection{Objectness Scoring \& Classification}

\noindent In this chapter, we introduce the Objectness Heatmap to identify sparse voxels \(x \in \mathcal{I}\) that are partially occupied by target objects \(\mathcal{O}\). Building upon the global context information decoded in the \(C\)-dimensional sparse voxel features \(f \in \mathbb{R}^{\mathcal{N}\times C}\), we use a fully sparse network module to aggregate fine geometric details and local context at reasonable resolution. With this staged heatmap strategy, we extract sharply segmented and high-resolution object segments to allow 6D pose estimation on a sample-level. For model training, we define a binary target function \(H_{obj}\left(\mathcal{I}, \mathcal{O}\right)=y \in \{ 0,1\}^{\mathcal{N}}\) that produces a binary target heatmap, with activations for voxels that are occupied \mbox{by \(\mathcal{O}\)}. For sparse voxels \(x_i \in \mathcal{I}\), we estimate per-voxel scores, which we denote as \(\hat{H}_{obj} \in [0,1]^{\mathcal{N}}\). We optimize the Objectness Heatmap using the Focal Loss \cite{focalloss} 
\begin{equation}
    \mathcal{L}_{obj}=\sum_{i=1}^\mathcal{N} FL(\hat{H}_{obj}\left(x_i\right), H_{obj}\left(x_i, \mathcal{O}\right)) \;\; ,
\end{equation}
\noindent as the cardinality of the background voxels still significantly outweighs that of foreground voxels. We extract the \(\mathcal{K}\) most potential sparse object voxels via \(topK\) selection. This further sparsifies the voxel grid to task-relevant object segments \(\mathcal{I}' =\{x_1,..,x_{\mathcal{K}}\} \subset \mathcal{I}\), while preserving fine geometric details and local context information. \\

\noindent We further predict per-voxel class estimates at this stage. We condition the classification head on objectness priors to overcome the class imbalance between fore- and background voxels. The heatmap features are projected with a linear layer and then applied as an additive bias to the classifier’s intermediate feature maps. We optimize the classifier using the weighted Cross-Entropy-Loss \(\mathcal{L}_{cls}\). 

\subsection{Sparse Transformer Block}
\begin{figure}[!t]
  \centering
   \includegraphics[width=1.\linewidth]{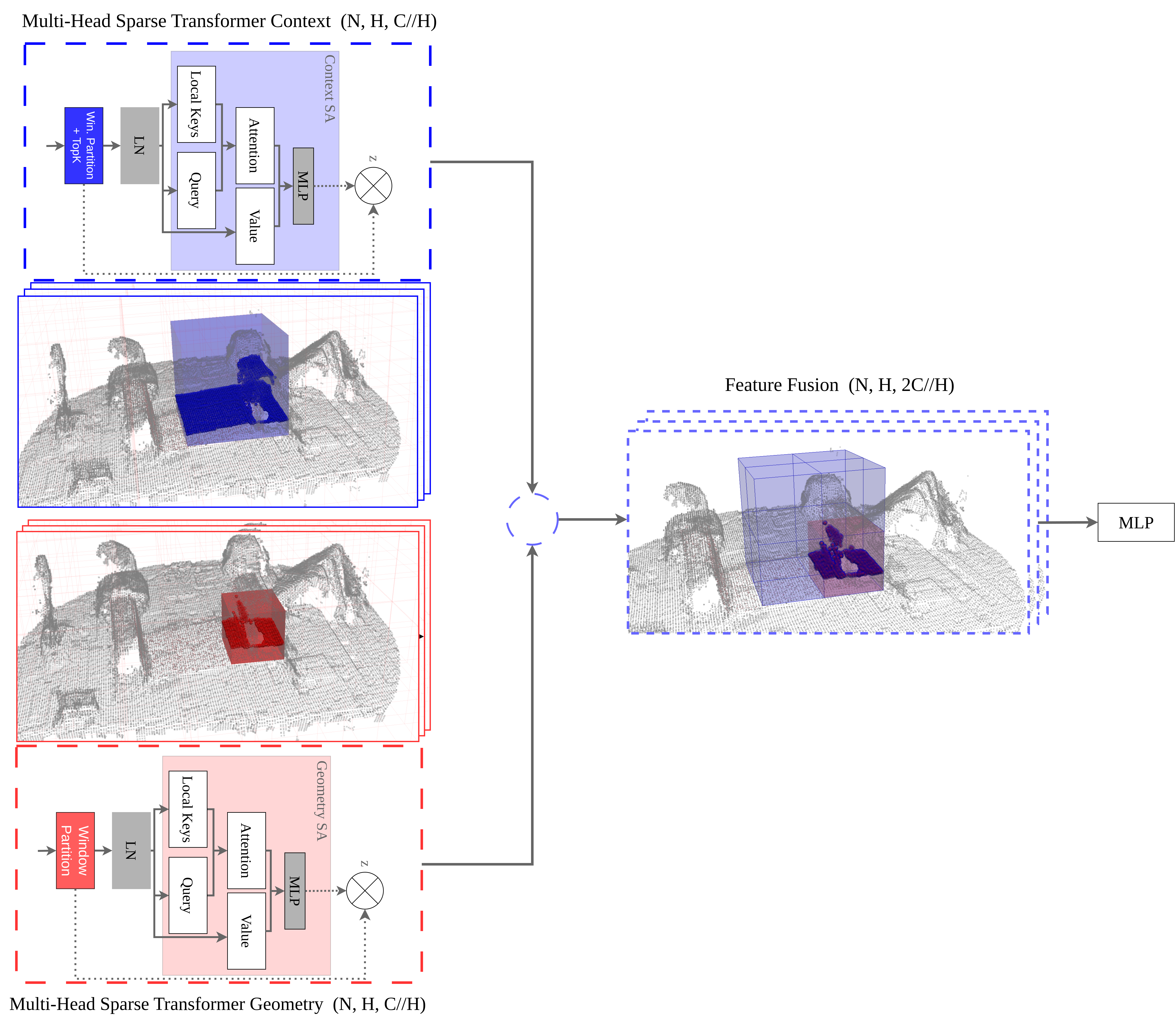}

   \caption{A single sparse transformer block. Multi-Head Window Self-Attention is performed in two branches with small and medium-sized windows. One attention branch captures fine geometric details, while the coarser accounts for neighborhood context. A MLP module enables dynamic feature selection.}
   \label{fig:onecol}
\end{figure}

\noindent 6D pose estimation in cluttered bin-picking scenes must frequently handle densely packed, similar parts that appear only in fragments due to occlusions and sensor noise. Accurate recovery of object poses requires both precise geometric detailing and an understanding of local context to disambiguate overlapping structures. Building on the sparse transformer backbone of \cite{sphereformer}, we introduce a sparse transformer block that dynamically focuses on the most informative voxel features, enabling the aggregation of both fine-grained geometry cues and neighborhood relations. \\ 

\noindent We perform multi-head self-attention in two branches with small and medium-sized windows. First, \(\mathcal{I}'\) is partitioned into non-overlapping cubic windows \(w\). The multi-head self-attention is conducted in each window independently as follows 
\begin{equation}
    \hat{q} = \text{MLP}_q\left(f\right), \; \hat{k} = \text{MLP}_k\left(f\right), \; \hat{v} = \text{MLP}_v\left(f\right) \;,
\end{equation}
\noindent where \(f \in \mathbb{R}^{\mathcal{K}_w\times C}\) are the \(C\)-dimensional input features within a window and \(\hat{q}, \hat{k}, \hat{v} \in \mathbb{R}^{\mathcal{K}_w \times C}\) denote the query, key and value, respectively. The number of input features \(\mathcal{K}_w\) is dynamic to account for the varying point density. The features are split into \(H\) attention heads such that \({q, k, v \in \mathbb{R}^{\mathcal{K}_w \times (H\times D)}}\), with \(C = D \; H\). We perform the dot product and weighted sum for head \(h\) as
\begin{equation}
    \begin{aligned}
        attn_h &= softmax\left(q_h \cdot k_h^T \right) \\
        z_h    &= attn_h \cdot v_h \;\; ,
    \end{aligned}
\end{equation}
\noindent with \(q_h, k_h, v_h \in \mathbb{R}^{\mathcal{K}_w\times D}\) denoting the features of the \(h\)-th head and the attention weight \(attn_h \in \mathbb{R}^{\mathcal{K}_w\times \mathcal{K}_w}\). The features of all heads are concatenated \(z = \text{concat}\left(\{z_0, .., z_{h-1}\}\right) \in \mathbb{R}^{\mathcal{K}_w \times C}\). A linear projection \(W \in \mathbb{R}^{C \times C}\) is applied that yields the self-attention features for a window, as \(\hat{z}_w = z \cdot W \;\; \in \mathbb{R}^{\mathcal{K}_w \times C}\). Considering all windows, we define the overall output as \(\hat{z} \in \mathbb{R}^{\mathcal{K}\times C}\). \\

\noindent We perform the multi-head window-attention on the small and medium-sized windows independently. We then concatenate \(\hat{z}_{small}\) and \(\hat{z}_{medium}\), which yields \(\hat{z} \in \mathbb{R}^{\mathcal{K}\times 2C}\). For context-adaptive feature selection, we propose a final linear layer block that projects \(\hat{z} \in \mathbb{R}^{\mathcal{K}\times 2C}\) to \(\mathbb{R}^{\mathcal{K} \times C}\). 

\subsection{6D Pose Estimation}
\noindent In this section, we introduce the 6D pose regression strategy. Our 6D pose estimation head operates over the entire scene jointly, predicting the poses of all target objects simultaneously through a novel per-voxel voting strategy. The head receives as input the sparsified voxel grid \(\mathcal{I}'\), and outputs the set of translation offset vectors \(\hat{t} \in \mathbb{R}^{\mathcal{K}\times 3}\) and rotations \(\hat{R} \in \mathbb{R}^{\mathcal{K}\times 6}\), as 6D-representation. It consists of a fully sparse network architecture, where sparse convolutional layers are interleaved with our custom sparse transformer blocks. Two parallel branches predict the per-voxel translation offsets \(\hat{t}\) and rotations \(\hat{R}\). The core of the head is the translation offset prediction mechanism: each sparse voxel \(x \in \mathcal{I}'\) that falls into a target object \(\mathcal{O}\) is assigned the relative translation offset to the object’s center. Let the translation target be denoted as \(t \in \mathbb{R}^{\mathcal{K}\times 3}\). We supervise the translation prediction using a smoothed L1-Loss \(\mathcal{L}_t(t,\hat{t})\). Similarly, for rotation estimation, each voxel associated with an object is assigned the corresponding ground truth rotation \(R \in \mathbb{R}^{\mathcal{K}\times 6}\). The rotation head is trained with a Chamfer Distance Loss \cite{pytorch3d} \(\mathcal{L}_{rot}(R, \hat{R}, \mathcal{O})\). During inference, the translation offset predictions naturally form spatial clusters (see Fig. \ref{fig:modelarch}, 5.1), which are grouped using \mbox{DBScan \cite{o3d}}. This clustering reliant mechanism enables the detection of an arbitrary number of objects without requiring predefined object counts or sizes. Given the cluster indices, we extract the most confidential predictions. Final pose estimates are obtained by averaging the most confident voxel-wise predictions within each cluster for both translation and rotation. We apply a final batched ICP \cite{pytorch3d}, to enhance alignment accuracy. \\

\noindent Overall, we train our network by minimizing the multi-task loss function
\begin{equation}
    \mathcal{L} = \lambda_1 \; \mathcal{L}_{RoI} + \lambda_2 \; \mathcal{L}_{obj} + \lambda_3 \; \mathcal{L}_{cls}+ \lambda_4 \; \mathcal{L}_{t} + \lambda_5 \; \mathcal{L}_{rot} \;\; ,
\end{equation}
where \(\lambda_1 = 1, \; \lambda_2 = 3, \; \lambda_3 = 2, \; \lambda_4 = 3, \; \lambda_5 = 1, \;\) are the weights for the individual loss functions.

\section{Experiments}
\subsection{Datasets}
\noindent We evaluate our approach on the recently published \mbox{IPD \cite{ipd}} and MV-YCB-SymMovCam \cite{symfm6d} multi-view datasets. The Industrial Plenoptic Dataset (IPD) is designed for industrial-grade 6D pose estimation in cluttered bin-picking scenarios, featuring 1,232 physical scenes with 10 objects captured in realistic multi-object, multi-instance settings. IPD supports multi-view pose estimation by providing object views from 13 multi-modal cameras. The MV-YCB-SymMovCam is a synthetic multi-view dataset of textured household objects from the YCB set. It comprises 33,332 annotated RGB-D images of 8,333 cluttered scenes, each captured from four distinct camera viewpoints placed in different quadrants around the scene. \\

\begin{table}[t]
\scriptsize
\centering
\caption{MV-YCB SymMovCam test results using 3 views with Point Cloud and TSDF input at (1) \(\vartheta=2 \text{ mm}\), (2) \(\vartheta=1.25 \text{ mm}\).}
\setlength{\tabcolsep}{3pt} 
\begin{tabular}{|c|c|c|c|c|}
\hline
\textbf{Method} & \textbf{ADD-S AUC} & \textbf{ADD-(S) AUC} & \textbf{ADD-S \(<\) 2} & \textbf{ADD-(S)\(<\) 2}  \\
\hline
\hline
PVN3D \cite{pvn3d}     & 75.0 & 68.5 & 77.2 & 64.5  \\[0.15ex]
FFB6D \cite{ffb6d}     & 79.9 & 75.6 & 81.1 & 74.5  \\[0.15ex]
MV6D \cite{mv6d}     & 92.8 & 88.7 & 96.3 & 91.6  \\[0.15ex]
SyMFM6D \cite{symfm6d}     & 94.2 & 91.6 & 96.6 & 93.6  \\[0.15ex]
\textbf{Ours (PC)} (1)         & 87.8 & 87.2 & 91.9 & 88.4  \\[0.15ex]
\textbf{Ours (TSDF)} (1)            & 90.1 & 88.5 & 93.1 & 90.2  \\[0.15ex]
\textbf{Ours (PC)} (2)         & 88.2 & 87.5 & 92.1 & 89.0  \\[0.15ex]
\textbf{Ours (TSDF)} (2)           & 90.6 & 89.1 & 93.5 & 90.8  \\[0.15ex]
\hline
\end{tabular}
\label{tab:mvycb-adds}
\end{table}

\begin{table}[]
\scriptsize
\centering
\caption{IPD Validation Results for TSDF (upper) and raw Point Cloud (bottom) input at \(\vartheta=1.25\text{ mm}\). }
\setlength{\tabcolsep}{2.5pt} 
\begin{tabular}{|c|c|c|c|c|c|c|c|c|c|}
\hline
\noalign{\vskip 0.2mm}
\textbf{Metrics} & \textbf{0} & \textbf{1} & \textbf{4} & \textbf{8} & \textbf{10} & \textbf{11} & \textbf{14} & \textbf{19} & \textbf{20}  \\
\hline
\noalign{\vskip 0.2mm}
MSPD     & 0.812 & 0.603 & 0.536 & 0.906 & 0.808 & 0.521 & 0.830 & 0.727 & 0.826  \\[0.25ex]
MSSD     & 0.622 & 0.538 & 0.482 & 0.789 & 0.715 & 0.431 & 0.699 & 0.691 & 0.764 \\[0.25ex]
MSSDmm   & 0.517 & 0.474 & 0.392 & 0.585 & 0.629 & 0.310 & 0.623 & 0.544 & 0.709  \\[0.25ex]
\hline
\hline
\noalign{\vskip 0.2mm}
\multicolumn{1}{|c|}{\textbf{AP}} & 0.683 & \textbf{AP25} & 0.637 & \textbf{AP25mm} & 0.531  & \multicolumn{4}{c|}{} \\
\hline
\noalign{\vskip 2mm}  
\hline
\noalign{\vskip 0.2mm}
\textbf{Metrics} & \textbf{0} & \textbf{1} & \textbf{4} & \textbf{8} & \textbf{10} & \textbf{11} & \textbf{14} & \textbf{19} & \textbf{20}  \\
\hline
\noalign{\vskip 0.2mm}
MSPD     & 0.782 & 0.608 & 0.496 & 0.753 & 0.864 & 0.492 & 0.833 & 0.687 & 0.822  \\[0.25ex]
MSSD     & 0.527 & 0.509 & 0.412 & 0.689 & 0.815 & 0.495 & 0.639 & 0.691 & 0.674 \\[0.25ex]
MSSDmm   & 0.487 & 0.466 & 0.341 & 0.597 & 0.617 & 0.291 & 0.607 & 0.539 & 0.712  \\[0.25ex]
\hline
\hline
\noalign{\vskip 0.2mm}
\multicolumn{1}{|c|}{\textbf{AP}} & 0.645 & \textbf{AP25} & 0.605 & \textbf{AP25mm} & 0.517  & \multicolumn{4}{c|}{} \\
\hline

\end{tabular}
\label{tab:ipd_results}
\end{table}

\subsection{Implementation Details}
\noindent We train our model for 20 epochs on the IPD dataset and 15 epochs on MV-YCB-SymMovCam using a single NVIDIA RTX 5090 GPU with 32 GB memory. For training stability, the RoI head is optimized separately for one warm-up epoch before all modules are trained jointly. Empirically, we set the truncation scalar \(\tau_{sdf} = 8 \; \vartheta_{tsdf}\). We set the standard deviations in Eq.(3), \(\sigma_c = 6\) and \(\sigma_b = 4\). For batched ICP refinement, we use the implementation of \cite{pytorch3d}. On the IPD validation set (\(\vartheta=2\text{mm}\)), we report a quantitative time analysis with mean runtimes of \(1.37\; \text{s}\) for sparse TSDF generation, \(0.78\;\text{s}\) for the network forward pass, and \(1.94\; \text{s}\) for post-processing, where mean runtimes are reported due to the scene-adaptive nature of our approach. For point cloud input, multi-view fusion, forward pass, and post-processing require \(1.18\; \text{s}\), \(0.72\; \text{s}\), and \(1.82\; \text{s}\), respectively. 

\subsection{Results \& Ablation Studies}
\noindent In this section, we conduct pilot experiments to evaluate the effectiveness of the proposed \textit{SDT-6D} framework. \\ 

\noindent For evaluation on the MV-YCB-SymMovCam dataset, we report the area-under-curve (AUC) for ADD(-S) as well as the precision metrics ADD(-S) \(< 2 \text{cm}\) and \mbox{ADD-S  \(< 2 \text{cm}\)}. Our method is compared against the established single-view baselines FFB6D\cite{ffb6d}, and PVN3D\cite{pvn3d}) as well as the multi-view methods MV6D\cite{mv6d}, and SyMFM6D\cite{symfm6d}. The results are shown in Tab. \ref{tab:mvycb-adds}. Single-view baseline results are taken from \cite{symfm6d}. Despite using depth data only, \textit{SDT-6D} achieves competitive results, significantly outperforming single-view methods and demonstrating the strength of multi-view fusion. We find that our depth-only method performs comparably to the multi-view and multi-modal approaches and exhibits scalability as accuracy improves with higher-resolution input representations. This shows that our staged heatmap strategy, by emphasizing task-relevant information, effectively compensates for the absence of color and appears to be a promising replacement for indiscriminate reduction strategies.  \\

\begin{table}[]
\scriptsize
\centering
\caption{IPD Validation Results for TSDF input at different resolution scales. The upper table reports results at \(\vartheta=2\text{mm}\), the lower table the results at  \(\vartheta=4\text{mm}\). }
\setlength{\tabcolsep}{2.5pt} 
\begin{tabular}{|c|c|c|c|c|c|c|c|c|c|}
\hline
\noalign{\vskip 0.2mm}
\textbf{Metrics} & \textbf{0} & \textbf{1} & \textbf{4} & \textbf{8} & \textbf{10} & \textbf{11} & \textbf{14} & \textbf{19} & \textbf{20}  \\
\hline
\noalign{\vskip 0.2mm}
MSPD     & 0.803 & 0.543 & 0.492 & 0.908 & 0.818 & 0.491 & 0.791 & 0.704 & 0.813  \\[0.25ex]
MSSD     & 0.601 & 0.515 & 0.434 & 0.793 & 0.769 & 0.411 & 0.637 & 0.647 & 0.755 \\[0.25ex]
MSSDmm   & 0.511 & 0.446 & 0.307 & 0.569 & 0.631 & 0.309 & 0.618 & 0.518 & 0.698  \\[0.25ex]
\hline
\hline
\noalign{\vskip 0.2mm}
\multicolumn{1}{|c|}{\textbf{AP}} & 0.663 & \textbf{AP25} & 0.618 & \textbf{AP25mm} & 0.512  & \multicolumn{4}{c|}{} \\
\hline
\noalign{\vskip 2mm}  
\hline
\noalign{\vskip 0.2mm}
\textbf{Metrics} & \textbf{0} & \textbf{1} & \textbf{4} & \textbf{8} & \textbf{10} & \textbf{11} & \textbf{14} & \textbf{19} & \textbf{20}  \\
\hline
\noalign{\vskip 0.2mm}
MSPD     & 0.782 & 0.502 & 0.458 & 0.886 & 0.803 & 0.467 & 0.769 & 0.694 & 0.783  \\[0.25ex]
MSSD     & 0.572 & 0.486 & 0.369 & 0.775 & 0.729 & 0.358 & 0.598 & 0.602 & 0.728 \\[0.25ex]
MSSDmm   & 0.481 & 0.429 & 0.217 & 0.537 & 0.591 & 0.266 & 0.572 & 0.486 & 0.663  \\[0.25ex]
\hline
\hline
\noalign{\vskip 0.2mm}
\multicolumn{1}{|c|}{\textbf{AP}} & 0.631 & \textbf{AP25} & 0.578 & \textbf{AP25mm} & 0.471  & \multicolumn{4}{c|}{} \\
\hline

\end{tabular}
\label{tab:scalability}
\end{table}


\noindent We further evaluate our method on the IPD dataset, focusing on the 6D pose estimation of industrial objects in highly cluttered bin-picking scenarios. To our knowledge, this is the first depth-only, multi-view approach applied to IPD. We report results on the IPD validation split in Tab. \ref{tab:ipd_results}, using the BOP \cite{bop} metrics: Mean Symmetric Point Distance (MSPD) and Mean Symmetric Surface Distance (MSSD). For both input representations, our model achieves reliable performance. This demonstrates the robustness of the staged heatmap architecture and its capacity to hierarchically isolate and refine task-relevant object regions from sparse 3D inputs, even in challenging industrial environments. The method performs particularly well in scenarios with partial visibility and inter-object occlusion, reflecting its suitability for practical bin-picking tasks. The results induce that leveraging the sparse TSDF representation enhances scene-level reasoning by providing smoother and more geometrically consistent surfaces, leading to improved performance. For reasons related to training stability, we excluded the T-Bracket (ID 18) from both training and evaluation (see Sec. 5 for a discussion). Our approach comprises a fully sparse architecture, which enables efficient processing at any resolution level. We evaluate scalability by varying the voxel resolution \(\vartheta\). In Tab. \ref{tab:scalability} we demonstrate, that our approach gracefully scales with resolution, offering a practical and scalable solution for real-world robotic bin picking tasks. We see that especially small objects with fine geometric details, e.g. the corner bracket (ID 4), necessitate a high resolution representation, as performance degrades considerably at coarser resolution. We further evaluate the impact of the proposed sparse transformer block in Tab. \ref{tab:transformer}. We remove the dual-branch design with the learnable fusion module and perform window self-attention for small and medium-sized windows solely. For fair comparison, we retrain each setting for 3 epochs. We observe that removing the dual-branch design leads to consistent performance degradation. The performance drop is especially pronounced when global context is excluded, indicating that window attention without adaptive fusion struggles to capture sufficient context due to its limited receptive field. These findings highlight the effectiveness of our transformer design in balancing geometric precision and contextual reasoning.

\begin{table}[]
\scriptsize
\centering
\caption{Ablation of the sparse transformer block by removing the dual-branch design and fusion module, using TSDF input at \(\vartheta=2\text{mm}\). }
\setlength{\tabcolsep}{2.5pt} 
\begin{tabular}{|c|c|c|c|c|}
\hline
\noalign{\vskip 0.2mm}
\textbf{} & \textbf{MSPD} & \textbf{MSSD} & \textbf{MSSDmm} & \textbf{AP} \\
\hline
\noalign{\vskip 0.2mm}
\textit{SDT-6D}     & 0.708 & 0.618 & 0.512 & 0.663 \\[0.25ex]
\textit{SDT-6D} w/o context branch    & 0.674 & 0.588 & 0.458 & 0.631 \\[0.25ex]
\textit{SDT-6D} w/o geometry branch   & 0.687 & 0.601 & 0.479 & 0.644 \\[0.25ex]
\hline

\end{tabular}
\label{tab:transformer}
\end{table}


\section{Limitations}
\noindent Our approach follows a depth-only strategy, which inherently demands objects that exhibit sufficient geometric expressiveness. Extremely low-profile objects, e.g. the \mbox{T-Bracket} in the IPD dataset, diminish under sensor noise and real-world artifacts, rendering them indistinguishable from background even at high resolution. We observe that including such geometrically degenerate objects degrades the performance of our staged heatmap approach and compromises holistic pose estimation. Consequently, we exclude this object from our pilot experiments on the IPD dataset. This limitation may be mitigated by integrating color information, which our framework can accommodate, and remains an avenue for future work.

\section{Conclusion}
\noindent We introduced \textit{SDT-6D}, a depth-only, multi-view framework for 6D pose estimation. At its core lies a staged heatmap strategy that imposes structured attention priors across the 3D scene, enabling precise focus on task-relevant foreground while suppressing background. This allows operation at high spatial resolution without compromising computational efficiency. Pilot experiments demonstrate competitive performance. Notably, the proposed staged heatmap mechanism yields fine‐grained, 3D object segments and seamlessly fuses holistic scene context via its encoded features. We believe that this can benefit recent depth-based detectors in overcoming assumptions for prenormalized detection toward full-scene understanding.
\newpage
{
    \small
    \bibliographystyle{ieeenat_fullname}
    \bibliography{main}
}

\end{document}